\documentclass[a4paper]{llncs}

 \usepackage[utf8]{inputenc}
 \usepackage[T1]{fontenc}

 \usepackage{underscore}
 \usepackage{svg}
 \usepackage{csquotes}
 \usepackage{hyperref}
 \usepackage{graphicx}
 \usepackage{mathtools}
 \usepackage{amssymb}
 \usepackage{colortbl}
 \usepackage{multirow}
 \usepackage{footnote}
 \usepackage[noend]{algpseudocode}
 \usepackage{cite}
 \usepackage{bbding}
 \usepackage{booktabs}
 \usepackage{todonotes}
 \usepackage{subfig}
 \usepackage[ruled,vlined,linesnumbered]{algorithm2e}
 \usepackage{hyperref}[hidelinks]
 \makesavenoteenv{tabular}
 \makesavenoteenv{table}
 \newcommand\alg[0]{CostNet}
 \begin{document}

 	\title{CostNet: An End-to-End Framework for Goal-Directed Reinforcement Learning}
 	\titlerunning{TODO}

	\author{Per-Arne Andersen\textsuperscript{\Envelope}\orcidID{0000-0002-7742-4907} 
	\and Morten Goodwin\orcidID{0000-0001-6331-702X}
	\and Ole-Christoffer Granmo\orcidID{0000-0002-7287-030X}}
	 \authorrunning{P. Andersen et al.}
	 
 	\institute{Department of ICT, University of Agder, Grimstad, Norway\\
 		\email{\{per.andersen,morten.goodwin,ole.granmo\}@uia.no}}
 	\maketitle              
	 \begin{abstract}
		Reinforcement Learning (RL) is a general framework concerned with an agent that seeks to maximize rewards in an environment. The learning typically happens through trial and error using explorative methods, such as \(\epsilon\)-greedy. There are two approaches, model-based and model-free reinforcement learning, that show concrete results in several disciplines. Model-based RL learns a model of the environment for learning the policy while model-free approaches are fully explorative and exploitative without considering the underlying environment dynamics. Model-free RL works conceptually well in simulated environments, and empirical evidence suggests that trial and error lead to a near-optimal behavior with enough training. On the other hand, model-based RL aims to be sample efficient, and studies show that it requires far less training in the real environment for learning a good policy.
        
		A significant challenge with RL is that it relies on a well-defined reward function to work well for complex environments and such a reward function is challenging to define. Goal-Directed RL is an alternative method that learns an intrinsic reward function with emphasis on a few explored trajectories that reveals the path to the goal state.
		
		This paper introduces a novel reinforcement learning algorithm for predicting the distance between two states in a Markov Decision Process. The learned distance function works as an intrinsic reward that fuels the agent's learning. Using the distance-metric as a reward, we show that the algorithm performs comparably to model-free RL while having significantly better sample-efficiently in several test environments.
 	\end{abstract}
 
 \pagestyle{empty}

 \setcounter{page}{1}
 \section{Introduction}
 Goal-directed reinforcement learning (GDRL) separates the learning into two phases, where phase one aims to solve the goal-directed exploration problem (GDE). To solve the GDE problem, the agent must determine at least one viable path from the initial state to the goal state. In phase two, the agent uses the learned path to find a near-optimal path. The two phases iterate until the agent policy is converged.

 Reinforcement learning (RL) classifies into two categories of algorithms. Model-free RL learns a policy or a value-function by interaction with the environment and succeeds in various simulated areas, including video-games \cite{Mnih2015, Silver2016a}, robotics \cite{Kober2013, Levine2015}, and autonomous vehicles \cite{Shah2018, Grigorescu2020}, but comes at the cost of efficiency. Specifically, model-free approaches suffer from low sample efficiency and are a fundamental limitation for application in real-world physical systems.
 
 On the other hand, Model-based reinforcement learning (MBRL) aims to learn a predictive model of the environment to increase sample efficiency. The agent samples from the learned predictive model, which reduces the required interaction with the environment. However, it is challenging to achieve good accuracy of the predictive model for many domains, specifically for high complexity environments. With high complexity comes high modeling error (model-bias) and it is perhaps the most common problem for unstable and collapsing policies in model-based RL. Recent work in model-based RL focuses primarily on learning high-dimensional and complex predictive models with graphics as part of the MDP. This complicates the model severely and limits long-horizon predictions as the prediction-error increases exponentially. 
 
 This paper address this issue with a combination of GDRL and MBRL by learning a predictive model and a distance model that describes the distance between two states. The learned predictive model abstracts the state-space to distance between state and goal, which reduce the state-complexity significantly. The learned distance is applied to the reward-function of Deep Q-Learning (DQN) \cite{Mnih2013} and accelerates the learning effectively. The proposed algorithm, \alg, is an end-to-end solution for goal-directed reinforcement learning where the main contributions are summarized as follows.
 \begin{enumerate}
	 \item \alg~for estimating the distance between arbitrary states and terminal states,
	 \item modified objective for DQN for efficient goal-directed reinforcement learning, and
	 \item the proposed method demonstrates excellent performance in simulated grid-like environments.
 \end{enumerate}

 The paper is organized as follows. Section~\ref{sec:background} details the preliminary work for the proposed method. Section~\ref{sec:literature-review} presents a detailed overview of related work. Section~\ref{sec:method} introduces \alg, a novel algorithm for cost-directed reinforcement learning. Section~\ref{sec:results} thoroughly presents the results of the proposed approach, and Section~\ref{sec:conclusion} summarizes the work and propose future work in Goal-Directed Reinforcement learning.
 
 \section{Background}
 \label{sec:background}
 Model-based reinforcement learning builds a model of the environment to derive its behavioral policy. The underlying mechanism is a Markov Decision Process (MDP), which mathematically defines the synergy between state, reward, and actions as a tuple \(M = (S, A, T, R)\), where \(S = \{s_n, \ldots, s_{t+n}\}\) is a set of possible states and \(A = \{a_n,\ldots,a_{t+n}\}\) is a set of possible actions. The state transition function  \( T : S \times A \times S \rightarrow [0, 1]\), which the predictive model tries to learn is a probability function such that \(T_{a_t}(s_t,s_{t+1})\) is the probability that current state \(s_t\) transitions to \(s_{t+1}\) given that the agent choses action \(a_t\). The reward function \(R : S \times A \rightarrow \mathbb{R}\) where \(R_{a_t}(s_t, s_{t+1})\) returns the immediate reward received on when taking action \(a\) in state \(s_t\) with transition to \(s_{t+1}\). The policy takes the form \(\pi = \{s_1, a_1, s_2, a_2, \ldots, s_n, a_n\}\) where \(\pi(a \vert s)\) denotes chosen action given a state. Model-based reinforcement learning divides primarily into three categories: 1) Dyna-based, 2) Policy Search-based, and 3) Shooting-based algorithms in which this work concerns Dyna-based approaches. The Dyna algorithm from \cite{Sutton1991} trains in two steps. First, the algorithm collects experience from interaction with the environment using a policy from a model-free algorithm (i.e., Q-learning). This experience is part of learning an estimated model of the environment, also referred to as a predictive model. Second, the agent policy samples imagined data generated by the predictive model and update its parameters towards optimal behavior.

Autoencoders are commonly used in supervised learning to encode arbitrary input to a compact representation, and using a decoder to reconstruct the original data from the encoding. The purpose of autoencoders is to store redundant data into a densely packed vector form. In its simplest form, an autoencoder consists of a feed-forward neural network where the input and output layer is of equal neuron capacity and the hidden layer smaller, used to compress the data. The model consists of an encoder \(Q(z \vert X)\), latent variable distribution \(P(z)\), and decoder \(P(\hat{X} \vert z)\). The input \(X\) is a vector that represents only a fraction of the ground truth. The objective is for the autoencoder to learn the distribution of all possible training samples, including data not in the training data, but nevertheless, part of the distribution \(P(X)\).  The final objective for the model is  \(\mathbb{E}[log P(X \vert z)] - D_{KL}[Q(z \vert X)  \| P(z)]\), where the first term denotes the reconstruction loss, similar to standard autoencoders and the second term the distance between the estimated latent-space and the ground truth space. The ground truth latent-space is difficult to define, and therefore it is assumed to be a Gaussian, and hence, the learned distribution should also be a Gaussian.

\section{Related Work}
\label{sec:literature-review}
Pioneering work of the goal-directed viewpoint of reinforcement learning, uniformly suggests that pre-processing of the state-representation (i.e., model-based RL) and careful reward modeling is the preferred method to perform efficient GDRL. The following section introduces related work in GDRL and relevant model-based reinforcement learning methods\footnote{The reader is referred to~\cite{Polydros2017} for an in-depth survey of MBRL-based methods.}. 

\subsection{Goal-Directed Reinforcement Learning}
Earlier studies have contributed significantly to improve the abilities to solve reinforcement learning problems with a goal-directed approach. Perhaps the most well-known study of the Goal-Directed Reinforcement Learning problem begins with Koenig and Simmons~\cite{Koenig1996}. Their approach splits the problem into two phases, known as Goal-directed exploration (GDEP) and knowledge exploitation. The study finds that the convergence of GDRL-based \(\hat{Q}\)- and Q-learning closely relates to the state representation and volume of prior knowledge. Furthermore, their work shows that computationally intractable problems are tractable with minor modifications to the state- representation. 

Braga and Araújo apply GDRL in~\cite{Braga1998}, using temporal-difference learning to collect prior knowledge and to create a reward and penalty surface explaining the environment dynamics. The map acts as an expert advisor for the TD algorithm and proves the policy performance. Their work shows that the concept of GDRL works well in grid-based environments and includes significantly better sample efficiency compared to Q-Learning. 

In~\cite{Matignon2006}, the authors study the importance of reward function and initial Q-values for GDRL. The authors thoroughly studied the effect of different initial states of the Q-table and found it challenging to design a generic algorithm for initially setting optimal parameters. However, they found that initial values impact the performance and sample efficiency considerably. Furthermore, the author shows that adding a \textit{goal bias leads to much faster learning} and recommends an \textit{adjustable continuous reward function}. More recently, Debnath et al. propose a hybrid approach, formalized as a GDRL problem, where the first phase optimizes a predictive model of the environment with samples from a model-free reinforcement learning policy. The second phase exploits the learned predictive model to improve the policy further, similar to \cite{Andersen2020}. The authors show that GDRL-based algorithms accelerate learning and improve sample efficiency considerably \cite{Debnath2018}.

\subsection{Model-Based Reinforcement Learning}

The Model-Ensemble Trust-Region Policy Optimization (ME-TRPO), formally proposed by \cite{Kurutach2018}, is a Dyna-based algorithm for learning a predictive model. The ME-TRPO method uses an ensemble of neural networks to form the predictive model, which significantly reduces model-bias, increasing its generalization abilities. The ensemble individually trains using single-step L2 loss in a supervised setting. After training of the algorithm, the authors use Trust-Region Policy Optimization from~ \cite{Schulman2015} as the model-free approach. The work shows significantly faster convergence in several continuous control tasks. 

The ME-TRPO method extends to Stochastic Lower Bound Optimization (SLBO)~\cite{Luo2018}. In comparison, SLBO modifies the single-step L2 loss to multi-step L2-norm loss to the train ensemble predictive model. The authors present a mathematical framework for the guaranteed monotonic improvement of the predictive model. 

In \cite{Janner2019}, the authors analyze previous methods and their capability to generalize well for longer time horizons. Their analysis suggests that the performance is good for shorter time horizons, but exponentially decrease as uncertainty appears when predicting longer rollouts. The proposed algorithm is called Model-based Policy Optimization (MBPO) and balance a trade-off between sample efficiency and performance. The paper suggests a prediction horizon between and 1-15 states, up to 200 states. In conclusion, MBPO shows that model-based approaches can outperform state-of-the-art model-free reinforcement learning when tuning appropriately.

\section{CostNet for Goal-Directed RL}
\label{sec:method}

CostNet is a combination of four disciplines in Deep Learning, 1) Goal Directed RL \cite{Koenig1996}, 2) Model-Based RL \cite{Sutton2018}, and 3) Variational Autoencoders \cite{Kingma2013} and forms a novel approach for learning the cost between states modeled after an MDP. The algorithm accumulates training data from using expert systems or random sampling. For systems where safety is a priority, it is advised to perform sampling according to manually defined risk constraints  at the cost of increased sample complexity \cite{Andersen2020}.

\begin{figure}
	\centering
   \includegraphics[width=1.0\linewidth]{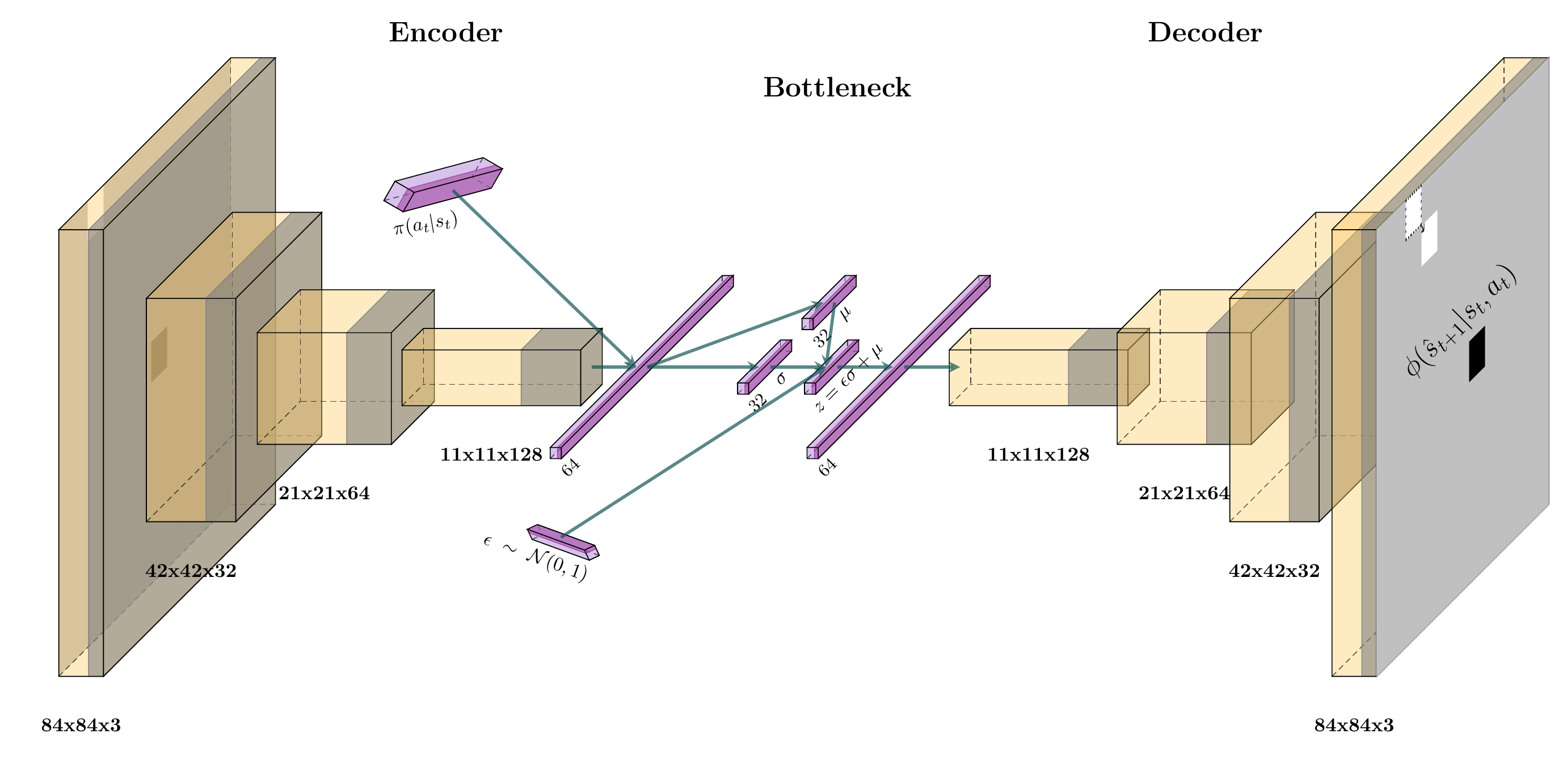}
   \caption{The encoder-latent-decoder architecture for learning a compact representation of states. The model is a convolutional variational autoencoder with three layers of convolutions before the latent-vector computation. The input is a state \(s_t\). The latent-space, \(z\) forms from an estimated \(\mu\), and \(\sigma\), mean and standard-deviation respectively, from a Gaussian. The \(\epsilon \sim N\) denotes sampling using the reparametrization trick, as described in \cite{Kingma2013}. On the right-hand side, the estimated latent-variable \(z\) reconstructs into the future state \(\hat{s}_{t+1}\)}
   \label{fig:architecture}
\end{figure}

The initial phase of training revolves around training a predictive model of the environment. Recent work indicates that state-of-the-art models suffer from sever policy drift after a few predictions \cite{Janner2019, Andersen2018b, Ha2018a}, and \alg~is no exception. Therefore, the problem is redefined to learning only the one-step prediction under a policy \(\phi(\hat{s}_{t+1} \vert s_t, a_t)^\pi\), where \(\phi\) denotes the predictive model. The predictive model is a variational autoencoder (VAE), where the goal is to map input (state) to latent-vectors that describes best possible describe the input. Figure \ref{fig:architecture}  illustrates the proposed structure for the encoder-latent-decoder model for CostNet. The input is an image of  an arbitrary state, and the hidden layers are convolutions with 32, 64, and 128 filters, a kernel size of 2, and a stride of 2 with ReLU activation, respectively. The latent-vector size is 64 neurons, but it is highly advised to fine-tune these hyper-parameters as the required embedding capacity varies on the state-complexity.

\begin{figure}
	\centering
	\includegraphics[width=1.0\linewidth]{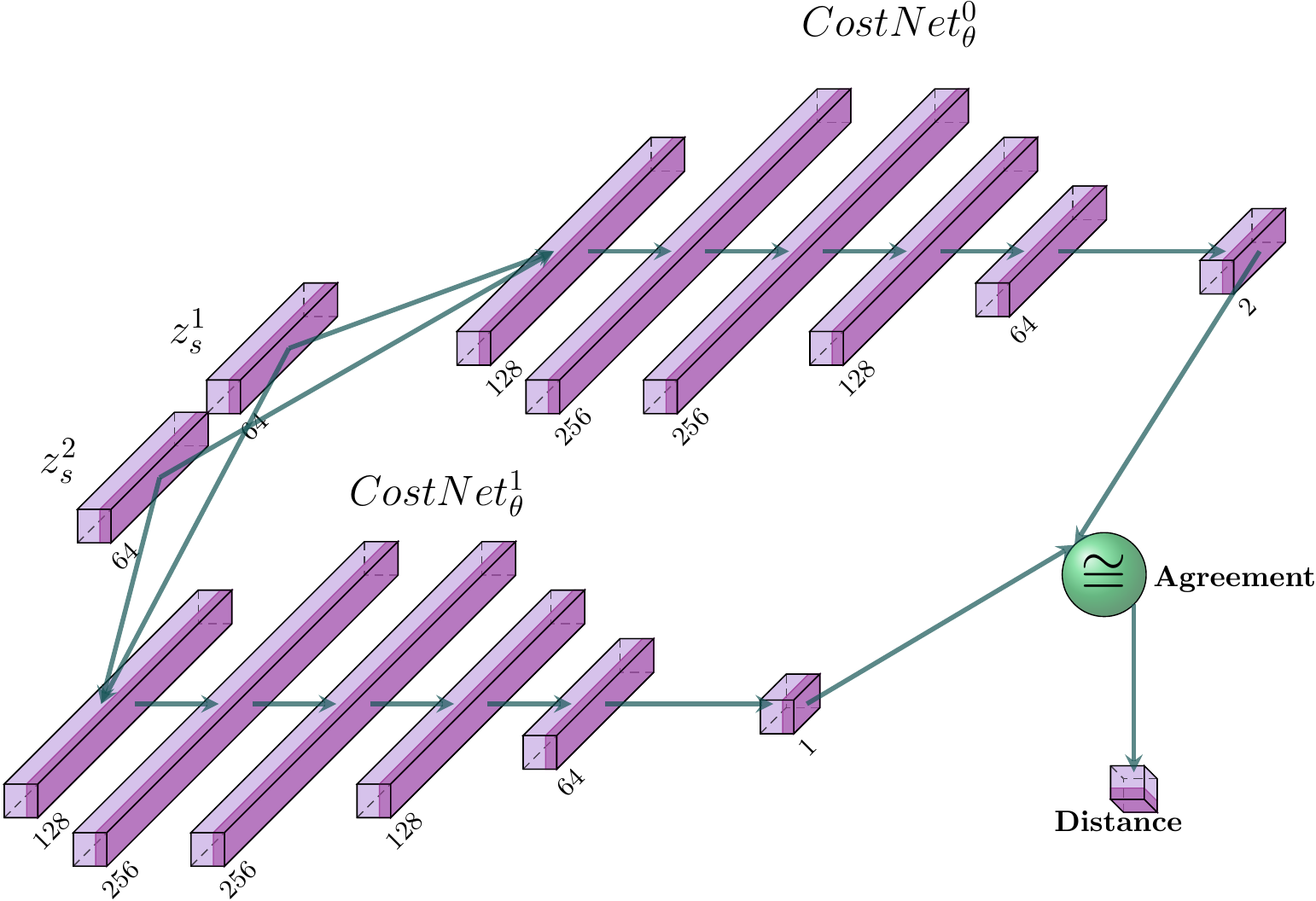}
	\caption{The proposed \alg~architecture. There are in-total two inputs, \(z_s^1\) and \(z_s^2\), that represent encoded states (see Figure \ref{fig:architecture}. The inputs are sent through two streams (models), \(\text{CostNet}_\theta^0\), and \(\text{CostNet}_\theta^1\), and learns two separate objectives using MSE. During training, both networks must agree on the answer for gradients to contribute in a positive direction. When both networks predict the same state to be closest to the goal state, the training is completed. The hidden layers are regular fully-connected with ReLU activation. The output for \(\text{CostNet}_\theta^0\)  activates with softmax, and \(\text{CostNet}_\theta^1\) with sigmoid activation.}
	\label{fig:costnet}
\end{figure}

The model-based approach to encode states as is developed as a method to improve the performance of the feed-forward neural network.
Figure~\ref{fig:costnet} shows the proposed architecture for the \alg~algorithm and consists of two models with different objectives. The first model, \(\text{CostNet}_\theta^0\), predicts which of the two states are closest to the goal, \(state_A\), or \(state_B\). The output is a vector that describes the probability of both \(state_A\) and \(state_B\) being closest to the goal. The second model, \(\text{CostNet}_\theta^1\), predicts the absolute distance to a goal state as a real number between 0 and 1, where 0 is at the goal state, and 1 is at maximum possible distance. Both networks train using mean squared error (MSE) loss, where the labels stem from the experience-buffer and the distance label from a backtracking algorithm. The predictions are considered correct (reliable) when there is an agreement between both networks, i.e. that \(\text{CostNet}_\theta^0\) correctly predicts which of \(state_A\) or \(state_B\) is closest, and \(\text{CostNet}_\theta^1\) predict the actual distance. 

To exemplify, consider the inputs \(z_s^1\) (\(state_A\)) and \(z_s^2\) (\(state_B\)) where \(z_s^1\)  is closest to the goal state. In this case, the first index in the vector from the \(\text{CostNet}_\theta^0\) prediction should be the largest signal, and the predicted distance from \(\text{CostNet}_\theta^1\) for \(z_s^1\) should be less for the similar prediction \(z_s^1\). If this is in place, we claim that the models are in agreement. When the agreement between the networks is consistent, the training is considered complete.

\begin{figure}
	\centering
	\includegraphics[width=0.66\linewidth]{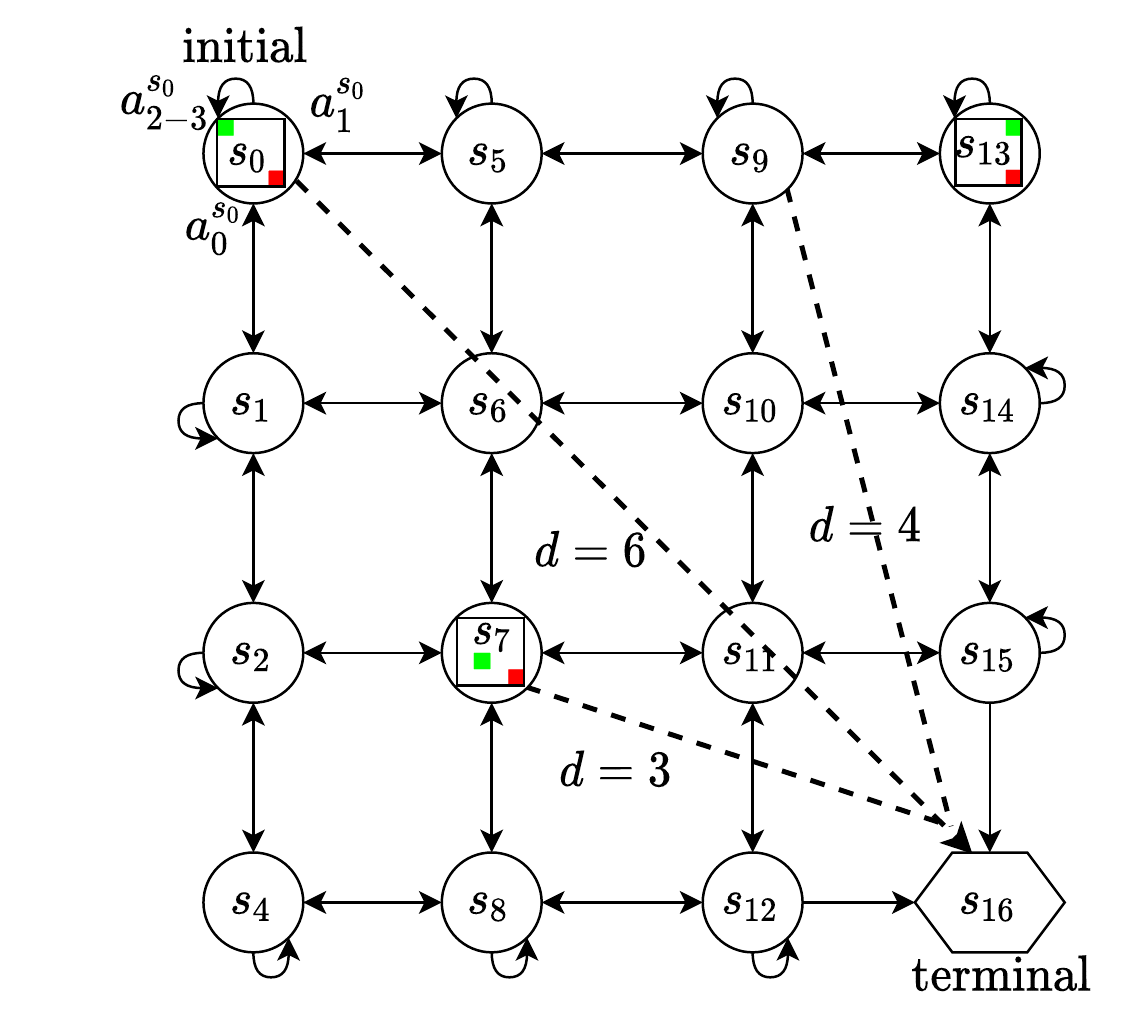}
	\caption{Illustration of the reformulation of the model-based MDP problem. State 0, 7, and 13 illustrate the usual state complexity for each node in the MDP graph; a pixel-based (full-state) representation. \alg, on the other hand, simplifies the state nodes to only a single metric: \textbf{distance}. This simplifies the complexity of the state-space significantly, and for GDRL-based approaches is sufficient representation.}
	\label{fig:mdp}
\end{figure}

Figure~\ref{fig:mdp} shows how the MDP is reduced to only focus on the distance to the goal-state. In regular MDP's the whole state information is represented at each node, illustrated by the inner square in state \(s_0, s_7, \text{and} s_{13}\). However, in this work, the MDP nodes only try to model the distance from one node to an arbitrary goal node. The problem with this formulation is that 1) there may be many goal states, and 2) agent must visit a goal state at least once. Therefore, the goal-directed approach works best in environments with less stochasticity in terms location of the goal states.

\begin{algorithm}[H]
	\SetAlgoLined
	\KwResult{Optimized policy \(\pi\) given a set of states \(S\), actions \(A\))}
	\textbf{Hyperparameters}:  Discount factor \(\gamma \in [0 \ldots 1]\), Learning-rate \(\alpha \in [0 \ldots 1]\), and Drift threshold \(\psi \in [0 \ldots 1]\)
	
	\textbf{Assumptions:} Experience-Replay (ER) from expert-system or random sampling, \(\Omega\)
	
	\While{training}{
		Train predictive model \(\phi(\hat{s}_{t+1}, z_t \vert s_t, s_{t+1}, a_t)\) from ER using objective \(\mathbb{E}[log P(X \vert z)] - D_{KL}[Q(z \vert X)  \| P(z)]\)

		\If{ \(\phi\) < \(\psi\)}{
			Train first supervised CostNet \(C_\theta^1\) using \(z_t, \hat{z}_{t+1}\) from \(\phi(z_t \vert s_t)\) and \(\phi(\hat{z}_{t+1} \vert \hat{s}_{t+1})\) with MSE loss.
		
			Train second supervised CostNet \(C_\theta^2\) using \(z_t\) from \(\phi(z_t \vert s_t)\) with MSE loss.
			
			\If{\(C_\theta^1 \cong C_\theta^2\) for most predictions (agreement)}{
				training = false
			}
		
		}
	}

	\While{training DQN}{
		Choose action \(a\) based on \(\epsilon\)-greedy
		Execute \(a\) at state \(s\) and get \(s_{t+1}, r\)
		Perform Q-Update:
		
		\(L_i(\theta_i) = \mathbb{E}_{s, a, r, s'\sim \rho(.)} \left[ (y_i - Q(s, a; \theta_i))^2 \right]\) where \(y_i =  \frac{r}{1 - C_\theta^2}+  \gamma \max_{a'} Q(s', a'; \theta_{i-1})\)
	}
	\caption{CostNet with Deep Q-Learning}
\end{algorithm}

\textbf{Predictive model}. The algorithm is summarized in the following line-by-line procedure. (Line 1) Initialize hyper-parameters for \(\gamma\), \(\alpha\), and \(\psi\) where the drift-threshold evaluates for \(n\) future predictions. When the algorithm is consistently below the threshold, training is complete. (Line 4) Train the predictive model using the ER-buffer using the objective from~\cite{Kingma2013} where \(Q(z \vert X)\) is the encoder to latent-space, \(P(z)\) is the distribution of latent-space, and \(P(X \vert z)\) is the decoder distribution. The object splits into two terms. The first term is the reconstruction loss (MSE), and the second term computes the KL distance between the predicted latent distribution \(Q(z \vert X)\) and the assumed normal distributed space \(P(z) \sim N(0, 1)\). (Line 5) when the predicted model is below the drift-threshold \(\psi\), the training concludes. 

\textbf{\alg}. (Line 6, Line 7) The \(\text{CostNet}_\theta^0\) model trains with encoded states \(z_t\) and \(z_{t+1}\) as input and produces a vector that predicts a probability for each of the states being closest to the goal state. A second model, \(\text{CostNet}_\theta^1\) predicts the distance for a single state, and when the predictions align consistently, the training concludes\footnote{ \(\text{CostNet}_\theta^x\)  where \(\theta\) is the parameters that are optimized for the model.}. (Line 13) The training CostNet is complete, and regular model-free RL is performed, using DQN \cite{Mnih2013} respectively. 

\textbf{Model-free RL}. (Line 14, Line 15) The agent performs regular sampling to accumulate ER for training. The training is performed similarly to \cite{Mnih2013}, but modifies the reward signal to account for the distance from the goal state. When the distance signal is weak, the agent receives little reward and otherwise large rewards for states close to the goal state.

\section{Results and Discussion}

\label{sec:results}
The \alg~algorithm is tested in four environments, CartPole-v1 from \cite{Brockman2016a}, DeepRTS GoldCollect~\cite{Andersen2018a}\footnote{The DeepRTS environment is available at: \url{https://github.com/cair/deep-rts}}, and DeepMaze StaticNoWalls \cite{Andersen2018b}\footnote{The DeepMaze environment is available at: \url{https://github.com/cair/deep-maze}}. The experiments compare \alg~to DQN~\cite{Mnih2013}, and PPO~ \cite{Schulman2017} for 1000000 timesteps during 100 experiments for statistical analysis of the results\footnote{The experiments are available here \url{https://github.com/cair/CostNet}}.
\subsection{Results}
\begin{table}[]
	\centering
	\caption{Hyper-parameters of \alg~algorithm}
	\label{tab:hyper-parameters}
	\begin{tabular}{@{}ll@{}}
		\toprule
		\textbf{Parameter}       & \textbf{Value} \\ \midrule
		Learning Rate (DQN)      & 0.01           \\
		Discount Factor (DQN)    & 0.95           \\
		ER-Size (DQN)            & 5000           \\
		Optimizer                & Adam           \\
		Optimizer Learning Rate  & 0.001          \\
		Drift-Threshold \(\psi\) & 0.3            \\ \bottomrule
	\end{tabular}
	\end{table}

\begin{figure*}
	\centering
	\subfloat[\label{fig:cartpole}]{%
		\setlength{\fboxsep}{0pt}\fbox{\includegraphics[width=0.32\textwidth]{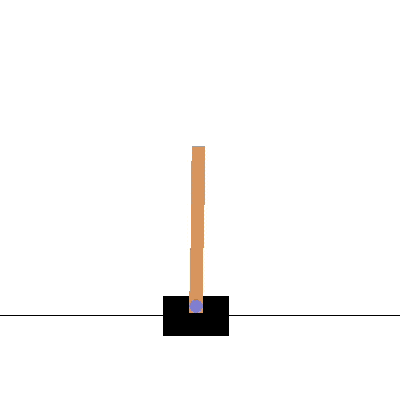}}}
	\hspace{\fill}
	\subfloat[\label{fig:deeprts} ]{%
		\setlength{\fboxsep}{0pt}\fbox{\includegraphics[width=0.32\textwidth]{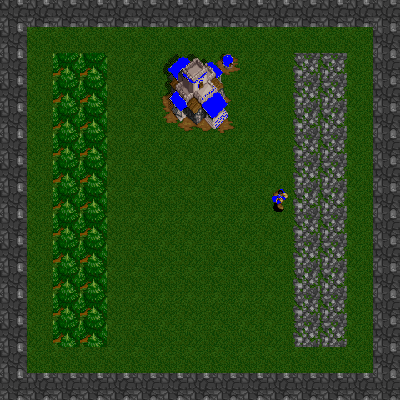}}}
	\hspace{\fill}
	\subfloat[\label{fig:deepmaze}]{%
		\setlength{\fboxsep}{0pt}\fbox{\includegraphics[, width=0.32\textwidth]{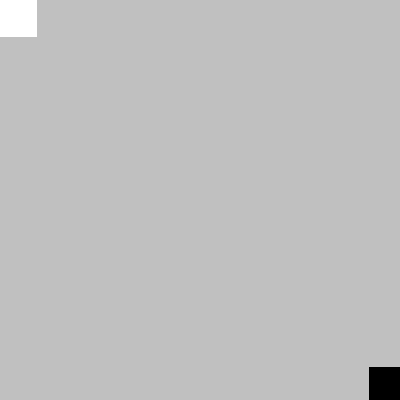}}}\\
	\caption{\label{fig:environments}Illustration of the experiments. (a) Cart Pole. Goal is to balance the pole until terminal state occurs. (b) DeepRTS GoldCollect. Gather as much gold as possible in a time-frame of 5 minutes. (c) DeepMaze. The player (white) must enter the terminal state (black) in shortest possible time.}
\end{figure*}

The hyper-parameters for \alg~are shown in Table~\ref{tab:hyper-parameters}. Figure \ref{fig:environments} shows the environments used in the experiments. The first environment is CartPole, a common benchmark for exploratory reinforcement learning research. The objective is to balance a pole on a cart for 500 timesteps at which the episodes end. The second environment is DeepRTS Gold-Collect, a simple environment where the goal is to accumulate as much gold as possible for 5 minutes. The optimal episodic reward for this environment is 1000. Finally, the DeepMaze StaticNoWalls environment is an \(11\times11\) grid structure where the goal is located at a fixed position. The reward for DeepMaze is the length of the maze because the agent and goal are located at opposite corners.

\begin{figure}
	\centering
	\includegraphics[width=1.0\linewidth]{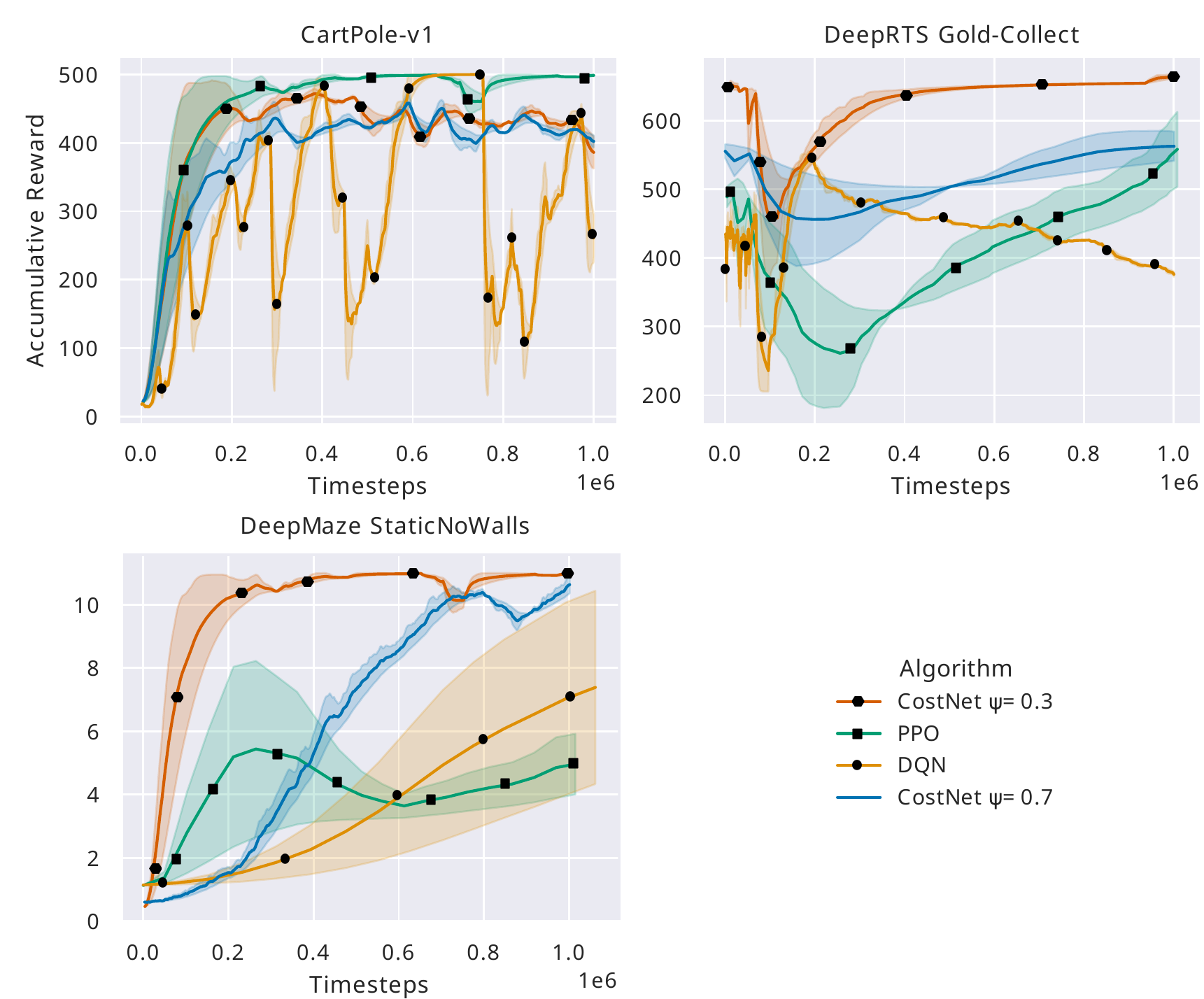}
	\caption{A comparison of PPO (square), DQN (circle), CostNet (unmarked and diamond) performance in CartPole, DeepRTS ,and Deep Maze environment. The y-axis shows the accumulated reward, and the x-axis is at which timestep. Every experiment runs for 100 episodes for 1 million timesteps. \alg~shows outstanding performance compared to fully model-free variants in two of three environments. In CartPole, PPO is superior, but \alg~closely follows. The experiments show that increasing the drift-threshold \(\psi\) also decreases performance and are an indication that CostNet impacts performance positively.}
	\label{fig:average-reward}
\end{figure}

Figure \ref{fig:average-reward} compares the performance of \alg~against two competing algorithms, DQN and PPO. Several parameters for the drift-threshold parameter\(\psi\) is tested, but a value of 0.3 seems to be stable across several environments. The PPO algorithm uses the parameters defined in \cite{Schulman2017} and for DQN in \cite{Mnih2015}. \alg~shows significantly better performance across all environments in terms of variance, seen clearly in the DeepMaze environment results. The primary reason for this is that the algorithm starts with a relatively good idea of the underlying environment dynamics from learning the predictive model. Furthermore, in terms of raw performance, the \alg~agent starts at near-optimal performance in some environments, such as the DeepRTS Gold-Collect environment. There are still challenges to be investigated, such as preventing divergence if the policy is already doing good behavior. Another problem is that \alg~demands initial data from expert systems which, is not possible in all environments. Regardless of these challenges, the algorithm is a good leap in the right direction and clearly ~\alg with a modified DQN reward-function, significantly increases the agent's performance, especially in more complex environments such as Deep RTS.

\subsection{Discussion}
\label{sec:discussion}
This paper's contribution shows that it is possible to learn distances between states in an MDP reliably and that the learned distance is useful for generic reward functions. The significance of applicability for \alg~spans across several disciplines. 
\textbf{Games} are perhaps the most obvious application for the algorithm as it is not always trivial to design reward functions generic enough to describe every state a complex MDP. While the proposed algorithm also suffers from generalization for multi-objective environments, it is still more accurate in learning reward functions compared to manually crafted functions. \textbf{Industry} the \alg~algorithm is applicable to the industry, especially in areas where the goal is stationary for all timesteps. One example is grid-warehousing, where agents operate on an A-to-B objective. However, upholding safety is a big concern when using RL-based algorithms, and therefore, GLDR-based approaches should be used with care.
 
\section{Future Work and Conclusion }
\label{sec:conclusion}

One question that merits future investigation is how to define optimal encode the state-space into latent-vectors. Using VAE is efficient, but still suffers from severe policy-shift for many-step future predictions. The proposed method is generic and should yield significant benefits from encoders that surpass VAE. Specifically, the VQ-VAE2 \cite{Razavi2019} shows promise, surpassing VAE in several disciplines. It would be interesting to see the effect VQ-VAE's discrete latent representation has on the overall performance when calculating state-to-state distances.

Another enticing direction for future work is analytical work for the \alg~architecture. The algorithm shows promising results empirically, which is often the case for deep reinforcement learning, but it remains future work to analytically prove the algorithm. Furthermore, in the extension of this work, the goal is to test the algorithm in many environments such as the MuJoCo, Atari Arcade, and DeepMind Lab environment to investigate its capabilities to generalize.

\alg~is a novel architecture for accelerating model-free reinforcement learning by combining goal-directed reinforcement learning and model-based reinforcement learning. The hybrid approach learns a predictive model, similar to \cite{Hafner2018}, but learns a simpler model, \alg, which captures only the distance between any given state and a terminal state. The algorithm outperforms DQN and PPO in several environments and shows outstanding stability during learning. Furthermore, \alg~shows promise for several disciplines, including games, industry, and autonomous driving. The hope is that future studies will lead to many more successes.

\bibliographystyle{splncs04}

\end{document}